%% file: main.tex
\documentclass{article}

\usepackage{amsmath}
\usepackage{amsfonts}
\usepackage{microtype}
\usepackage{graphicx}
\usepackage{subfigure}
\usepackage{booktabs} % for professional tables
\usepackage{mathtools}
\usepackage{bm}

\usepackage{hyperref}

\usepackage[accepted]{icml2020}

\DeclareMathOperator*{\argmax}{arg\,max}

\DeclareMathOperator{\diag}{Diag}

\DeclarePairedDelimiterX{\infdivx}[2]{(}{)}{%
  #1\;\delimsize\|\;#2%
}
\newcommand{\infdiv}{D_{KL}\infdivx}

\icmltitlerunning{Source Separation with Deep Generative Priors}

\begin{document}

\twocolumn[
\icmltitle{Source Separation with Deep Generative Priors}



\icmlsetsymbol{equal}{*}

\begin{icmlauthorlist}
\icmlauthor{Vivek Jayaram}{equal}
\icmlauthor{John Thickstun}{equal}
\end{icmlauthorlist}

\icmlcorrespondingauthor{Vivek Jayaram}{{\tt vjayaram@cs.washington.edu}}
\icmlcorrespondingauthor{John Thickstun}{{\tt thickstn@cs.washington.edu}}

\icmlkeywords{Machine Learning, ICML}

\vskip 0.3in
]



\printAffiliationsAndNotice{\textsuperscript{*} Equal contribution. Paul G. Allen School of Computer Science and Engineering, University of Washington}  % otherwise use the standard text.

\begin{abstract}
Despite substantial progress in signal source separation, results for richly structured
data continue to contain perceptible artifacts. In contrast, recent deep generative
models can produce authentic samples in a variety of domains that are indistinguishable
from samples of the data distribution.
This paper introduces a Bayesian approach to source separation that uses generative
models as priors over the components of a mixture of sources, and noise-annealed Langevin dynamics
to sample from the posterior distribution of sources given a mixture. 
This decouples the source separation problem from generative modeling, enabling
us to directly use cutting-edge generative models as priors.
The method achieves state-of-the-art performance for MNIST digit separation. We introduce
new methodology for evaluating separation quality on richer datasets, providing quantitative
evaluation of separation results on CIFAR-10. We also provide qualitative results on LSUN.
\end{abstract}

\input{intro.tex}
\input{related.tex}
\input{langevin.tex}
\input{eval.tex}
\input{separation.tex}

\input{conclude.tex}

\section*{Acknowledgements}
We thank Zaid Harchaoui, Sham M. Kakade, Steven Seitz, and Ira Kemelmacher-Shlizerman for valuable discussion and computing resources. This work was supported by the National Science Foundation Grant DGE-1256082.

\clearpage
\bibliography{sourcesep}
\bibliographystyle{icml2020}

\clearpage

\appendix
\onecolumn

\input{appendix.tex}

\end{document}

%% file: intro.tex
\section{Introduction}\label{sec:intro}

The single-channel source separation problem \cite{davies2007source} asks us to decompose a mixed signal $\textbf{m} \in \mathcal{X}$ into a linear combination of $k$ components $\textbf{x}_1,\dots,\textbf{x}_k \in \mathcal{X}$ with scalar mixing coefficients $\alpha_i \in \mathbb{R}$:
\begin{equation}\label{eqn:ss}
\textbf{m} = g(\textbf{x}) \equiv \sum_{i=1}^k \alpha_i\textbf{x}_i.
\end{equation}
This is motivated by, for example, the ``cocktail party problem'' of isolating the utterances of individual speakers $\textbf{x}_i$ from an audio mixture $\textbf{m}$ captured at a busy party, where multiple speakers are talking simultaneously.

With no further constraints or regularization, solving Equation \eqref{eqn:ss} for $\textbf{x}$ is highly underdetermined. Classical ``blind'' approaches to single-channel source separation resolve this ambiguity by privileging solutions to \eqref{eqn:ss} that satisfiy mathematical constraints on the components $\textbf{x}$, such as statistical independence \cite{davies2007source} sparsity \cite{lee1999blind} or non-negativity \cite{lee1999learning}. These constraints can be be viewed as weak priors on the structure of sources, but the approaches are blind in the sense that they do not require adaptation to a particular dataset.

Recently, most works have taken a data-driven approach. To separate a mixture of sources, it is natural to suppose that we have access to samples $\textbf{x}$ of individual sources, which can be used as a reference for what the source components of a mixture are supposed to look like. This data can be used to regularize solutions of Equation \eqref{eqn:ss} towards structurally plausible solutions. The prevailing way to do this is to construct a supervised regression model that maps an input mixture $\textbf{m}$ to components $\textbf{x}_i$ \cite{huang2014singing,halperin2018neural}. Paired training data $(\textbf{m},\textbf{x})$ can be constructed by summing randomly chosen samples from the component distributions $\textbf{x}_i$ and labeling these mixtures with the ground truth components.

\begin{figure*}[ht]
\centering
\hspace*{-2mm}\includegraphics[scale=0.79]{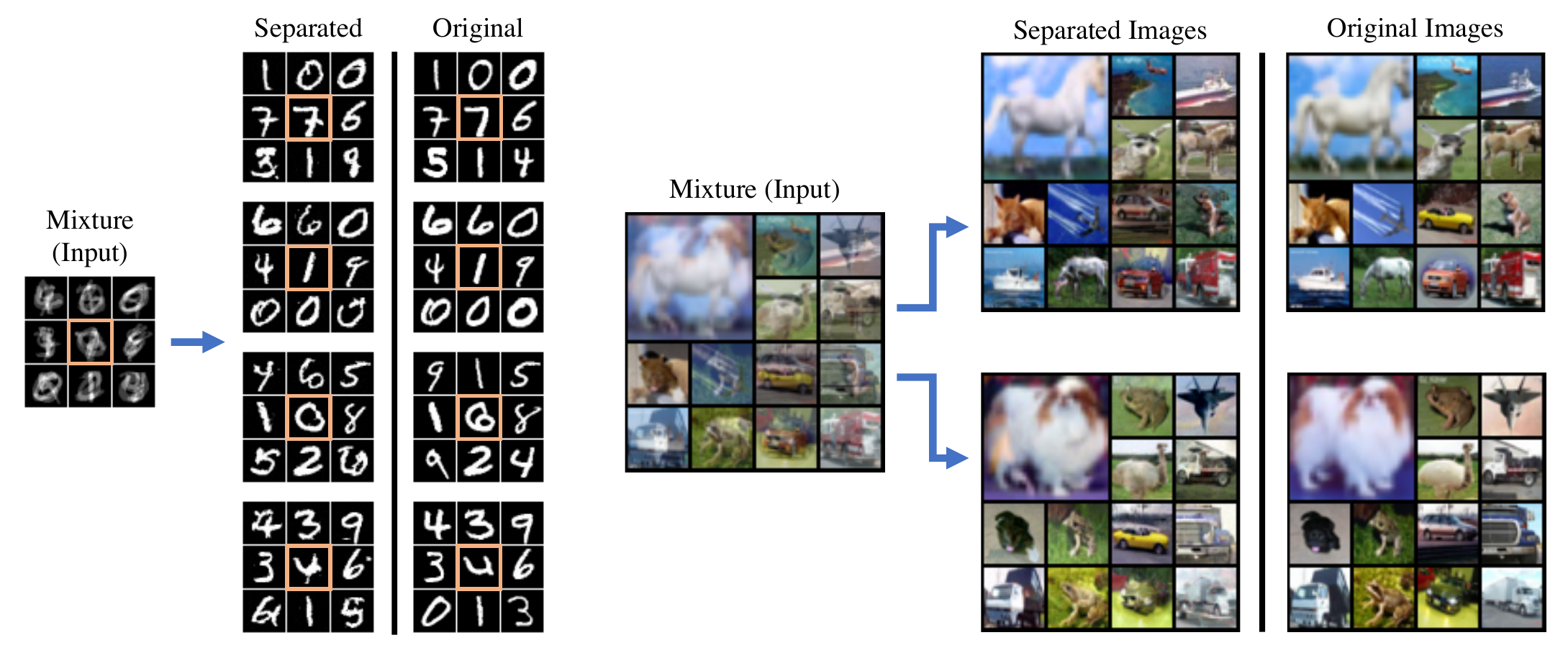}
\vspace{-6mm}
\caption{Separation results for mixtures of four images from the MNIST dataset (Left) and two images from the CIFAR-10 dataset (Right), using BASIS with the NCSN \cite{song2019generative} generative model as a prior over images. We draw attention to the central panel of the MNIST results (highlighted in orange), which shows how a mixture can be separated in multiple ways.}
\label{fig:teaser}
\end{figure*}

Instead of regressing against components $\textbf{x}$, we use samples to train a generative prior $p(\textbf{x})$; we separate a mixed signal $\textbf{m}$ by sampling from the posterior distribution $p(\textbf{x}|\textbf{m})$. For some mixtures this posterior is quite peaked, and sampling from $p(\textbf{x}|\textbf{m})$ recovers the only plausible separation of $\textbf{m}$ into likely components. But in many cases, mixtures are highly ambiguous: see, for example, the orange-highlighted MNIST images in Figure \ref{fig:teaser}. This motivates our interest in sampling, which explores the space of plausible separations. In Section \ref{sec:langevin} we introduce a procedure for sampling from the posterior, an extension of the noise-annealed Langevin dynamics introduced in \citet{song2019generative}, which we call Bayesian Annealed SIgnal Source separation: ``BASIS'' separation.

Ambiguous mixtures pose a challenge for traditional source separation metrics, which presume that the original mixture components are identifiable and compare the separated components to ground truth. For ambiguous mixtures of rich data, we argue that recovery of the original mixture components is not a well-posed problem. Instead, the problem we aim to solve is finding components of a mixture that are consistent with a particular data distribution. Motivated by this perspective, we discuss evaluation metrics in Section \ref{sec:eval}.

Formulating the source separation problem in a Bayesian framework decouples the problem of source generation from source separation. This allows us to leverage pre-trained, state-of-the-art, likelihood-based generative models as prior distributions, without requiring architectural modifications to adapt these models for source separation. Examples of source separation using noise-conditioned score networks (NCSN) \cite{song2019generative} as a prior are presented in Figure \ref{fig:teaser}. Further separation results using NCSN and Glow \cite{kingma2018glow} are presented in Section~\ref{sec:experiments}.

%% file: related.tex
\section{Related Work}\label{sec:rel}

\textbf{Blind separation}. Work on blind source separation is data-agnostic, relying on generic mathematical properties to privilege particular solutions to \eqref{eqn:ss} \cite{comon1994independent,bell1995information,davies2007source,huang2012singing}. Because blind methods have no access to sample components, they face the challenging task of modeling the distribution over unobserved components while simultaneously decomposing mixtures into likely components. It is difficult to fit a rich model to latent components, so blind methods often rely on simple models such as dictionaries to capture the structure of these components.

One promising recent work in the blind setting is Double-DIP \cite{gandelsman2019double}. This work leverages the unsupervised Deep Image Prior \cite{ulyanov2018deep} as a prior over signal components, similar to our use of a trained generative model. But the authors of this work document fundamental obstructions to applying their method to single-channel source separation; they propose using multiple image frames from a video, or multiple mixtures of the same components with different mixing coefficients $\alpha$. This multiple-mixture approach is common to much of the work on blind separation. In contrast, our approach is able to separate components from a single mixture.

\textbf{Supervised regression}. Regression models for source separation learn to predict components for a mixture using a dataset of mixed signals labeled with ground truth components. This approach has been extensively studied for separation of images \cite{halperin2018neural}, audio spectrograms \cite{huang2014singing,huang2015joint,nugraha2016multichannel,jansson2017singing}, and raw audio \cite{lluis2018end,stoller2018wave,defossez2019music}, as well as more exotic data domains, e.g. medical imaging \cite{nishida1999signal}. By learning to predict components (or equivalently, masks on a mixture) this approach implicitly builds a generative model of the signal components. This connection is made more explicit in recent work that uses GAN's to force components emitted by a regression model to match the distribution of a given dataset \cite{zhang2018single,stoller2018adversarial}. 

The supervised approach takes advantage of expressive deep models to capture a strong prior over signal components.
But it requires specialized model architectures trained specifically for the source separation task. In contrast, our approach leverages standard, pre-trained generative models for source separation. Furthermore, our approach can directly exploit ongoing advances in likelihood-based generative modeling to improve separation results.

\textbf{Signal Dictionaries}. Much work on source separation is based on the concept of a signal dictionary, most notably the line of work based on non-negative matrix factorization (NMF) \cite{lee2001algorithms}. These approaches model signals as combinations of elements in a latent dictionary. Decomposing a mixture into dictionary elements can be used for source separation by (1) clustering the elements of the dictionary and (2) reconstituting a source using elements of the decomposition associated with a particular cluster.

Dictionaries are typically learned from data of each source type and combined into a joint dictionary, clustered by source type \cite{schmidt2006single,virtanen2007monaural}. The blind setting has also been explored, where the clustering is obtained without labels by e.g. k-means \cite{spiertz2009source}. Recent work explores more expressive decomposition models, replacing the linear decompositions used in NMF with expressive neural autoencoders \cite{smaragdis2017neural,venkataramani2017neural}.

When the dictionary is learned with supervision from labeled sources, dictionary clusters can be interpreted as implicit priors on the distributions over components. Our approach makes these prior explicit, and works with generic priors that are not tied to the dictionary model. Furthermore, our method can separate mixed sources of the same type, whereas mixtures of sources with similar structure present a conceptual difficulty for dictionary-based methods.

\textbf{Generative adversarial separation}. Recent work by \citet{subakan2018generative} and \citet{Kong2019SingleChannelSS} explores the intriguing possibility of optimizing $\textbf{x}$ given a mixture $\textbf{m}$ to satisfy \eqref{eqn:ss}, where components $\textbf{x}_i$ are constrained to the manifold learned by a GAN. The GAN is pre-trained to model a distribution over components. Like our method, this approach leverages modern deep generative models in a way that decouples generation from source separation. We view this work as a natural analog to our likelihood-based approach in the GAN setting.

\textbf{Likelihood-based approaches}. Our approach is similar in spirit to older ideas based on maximum a posteriori estimation \cite{geman1984stochastic} likelihood maximization \cite{pearlmutter1997maximum,roweis2001one} and Bayesian source separation \cite{benaroya2005audio}. We build upon their insights, with the advantage of increased computational resources and modern expressive generative models.

%% file: langevin.tex
\section{BASIS Separation}\label{sec:langevin}

We consider the following generative model of a mixed signal \textbf{m}, relaxing the mixture constraint $g(\textbf{x}) = \textbf{m}$ to a soft Gaussian approximation:
\begin{align}
\textbf{x} &\sim p,\\
\textbf{m} &\sim \mathcal{N}\left(g(\textbf{x}),{}\gamma^2 I\right).
\end{align}
This defines a joint distribution $p_\gamma(\textbf{x},\textbf{m}) = p(\textbf{x})p_\gamma(\textbf{m}|\textbf{x})$ over signal components $\textbf{x}$ and mixtures $\textbf{m}$, and a corresponding posterior distribution
\begin{equation}\label{eqn:posterior}
p_\gamma(\textbf{x}|\textbf{m}) = p(\textbf{x})p_\gamma(\textbf{m}|\textbf{x})/p_\gamma(\textbf{m}).
\end{equation}
In the limit as $\gamma^2 \to 0$, we recover the hard constraint on the mixture $\textbf{m}$ given by Equation \eqref{eqn:ss}.

BASIS separation (Algorithm \ref{alg:sourcesep}) presents an approach to sampling from \eqref{eqn:posterior} based on the discussion in Sections \ref{sec:dynamics} and \ref{sec:acceleration}. In Section \ref{sec:gradients} we discuss the behavior of the gradients $\nabla_\textbf{x} \log p(\textbf{x})$, which motivates some of the hyper-parameter choices in Section \ref{sec:hyper}. We describe a procedure to construct the noisy models $p_{\sigma_i}$ required for BASIS in Section \ref{sec:models}.

\subsection{Langevin dynamics}\label{sec:dynamics}

\begin{algorithm}[tb]
   \caption{BASIS Separation}
   \label{alg:sourcesep}
\begin{algorithmic}
   \STATE {\bfseries Input:} $\textbf{m} \in \mathcal{X}$, $\{\sigma_i\}_{i=1}^L$, $\delta$, $T$
   \STATE Sample $\textbf{x}_1,\dots,\textbf{x}_k \sim \text{Uniform}(\mathcal{X})$
   \FOR{$i \gets 1$ {\bfseries to} $L$}
   \STATE $\eta_i \gets \delta \cdot \sigma_i^2/\sigma_L^2$
   \FOR{$t=1$ {\bfseries to} $T$}
       \STATE Sample $\varepsilon_t \sim \mathcal{N}(0,I)$
       \STATE $\textbf{u}^{(t)} \gets \textbf{x}^{(t)}+ \eta_i \nabla_\textbf{x} \log p_{\sigma_i}(\textbf{x}^{(t)}) + \sqrt{2\eta}\varepsilon_t$
       \STATE $\textbf{x}^{(t+1)} \gets \textbf{u}^{(t)} - \frac{\eta_i}{\sigma_i^2} \diag(\alpha)\left(\textbf{m} - g(\textbf{x}^{(t)})\right)$
   \ENDFOR
   \ENDFOR
\end{algorithmic}
\end{algorithm}
\setlength{\textfloatsep}{0.3cm}
\setlength{\floatsep}{0.3cm}

Sampling from the posterior distribution $p_\gamma(\textbf{x}|\textbf{m})$ looks formidable; just computing Equation \eqref{eqn:posterior} requires evaluation of the partition function $p_\gamma(\textbf{m})$. But using Langevin dynamics \cite{neal2011mcmc,welling2011bayesian} we can sample $\textbf{x} \sim p_\gamma(\cdot|\textbf{m})$ while avoiding explicit computation of $p_\gamma(\textbf{x}|\textbf{m})$. Let $\textbf{x}_0~\sim~\text{Uniform}(\mathcal{X})$, $\varepsilon_t \sim \mathcal{N}(0,I)$, and define a sequence
\begin{align}\label{eqn:langevin}
&\textbf{x}^{(t+1)} \equiv \textbf{x}^{(t)} \hspace*{-.5mm}+\hspace*{-.5mm} \eta\nabla_\textbf{x} \log p_\gamma(\textbf{x}^{(t)}|\textbf{m}) \hspace*{-.5mm}+\hspace*{-.5mm} \sqrt{2\eta}\varepsilon_t\\
&= \textbf{x}^{(t)} \hspace*{-.5mm}+\hspace*{-.5mm} \eta \nabla_\textbf{x} \left(\log p(\textbf{x}^{(t)}) \hspace*{-.5mm}+\hspace*{-.5mm} \tfrac{1}{2\gamma^2} \|\textbf{m} \hspace*{-.5mm}-\hspace*{-.5mm} g(\textbf{x}^{(t)})\|^2\right) \hspace*{-.5mm}+\hspace*{-.5mm} \sqrt{2\eta}\varepsilon_t.\notag
\end{align}
Observe that $\nabla_\textbf{x} \log p_\gamma(\textbf{m}) = 0$, so this term is not required to compute~\eqref{eqn:langevin}. By standard analysis of Langevin dynamics, as the step size $\eta \to 0$, $\lim_{t \to \infty} \infdiv{\textbf{x}_t}{\textbf{x}|\textbf{m}} = 0$, under regularity conditions on the distribution $p_\gamma(\textbf{x}|\textbf{m})$.

If the prior $p(\textbf{x})$ is parameterized by a neural model, then gradients $\nabla_\textbf{x} \log p(\textbf{x})$ can be computed by automatic differentiation with respect to the inputs of the generator network. This family of likelihood-based models includes autoregressive models \cite{salimans2017pixelcnn++,parmar2018image}, the variational autoencoder \cite{kingma2013auto,van2017neural}, or flow-based models \cite{dinh2016density,kingma2018glow}. Alternatively, if gradients of the distribution are modeled \cite{song2019generative}, then $\nabla_\textbf{x}\log p(\textbf{x})$ can be used directly.

\subsection{Accelerated mixing}\label{sec:acceleration}

To accelerate mixing of \eqref{eqn:langevin} we adopt a simulated annealing schedule over noisy approximations to the model $p(\textbf{x})$, extending the unconditional sampling algorithm proposed in \citet{song2019generative} to accelerate sampling from the posterior distribution $p_\gamma(\textbf{x}|\textbf{m})$. Let $p_\sigma(\textbf{x})$ denote the distribution of $\textbf{x} + \epsilon_\sigma$ for $\textbf{x} \sim p$ and $\epsilon_\sigma \sim \mathcal{N}(0,\sigma^2I)$. We define the noisy joint likelihood $p_{\sigma,\gamma}(\textbf{x},\textbf{m}) \equiv p_\sigma(\textbf{x})p_\gamma(\textbf{m}|\textbf{x})$, which induces a noisy posterior approximation $p_{\sigma,\gamma}(\textbf{x}|\textbf{m})$. At high noise levels $\sigma$, $p_\sigma(\textbf{x})$ is approximately Gaussian and irreducible, so the Langevin dynamics \eqref{eqn:langevin} will mix quickly. And as $\sigma \to 0$, $\infdiv{p_\sigma}{p} \to 0$. This motivates defining the modified Langevin dynamics
\begin{align}\label{eqn:langevin_fast}
\textbf{x}^{(t+1)} &\equiv \textbf{x}^{(t)} + \eta\nabla_\textbf{x} \log p_{\sigma,\gamma}(\textbf{x}^{(t)}|\textbf{m}) + \sqrt{2\eta}\varepsilon_t.
\end{align}
The dynamics \eqref{eqn:langevin_fast} approximate samples from $p(\textbf{x}|g(\textbf{x}) = \textbf{m})$ as $\eta \to 0$, $\gamma^2 \to 0$, $\sigma^2 \to 0$, and $t \to \infty$. An implementation of these dynamics, annealing $\eta$, $\gamma^2$, and $\sigma^2$ as $t \to \infty$ according to the hyper-parameter settings presented in Section \ref{sec:hyper}, is presented in Algorithm \ref{alg:sourcesep}.

We anneal $\eta$, $\gamma^2$, and $\sigma^2$ using a heuristic introduced in \citet{song2019generative}: the idea is to maintain a constant signal-to-noise ratio (SNR) between the expected size of the posterior log-likelihood gradient term $\eta\nabla_\textbf{x} \log p_{\sigma,\gamma}(\textbf{x}|\textbf{m})$ and the expected size of the Langevin noise $\sqrt{2\eta}\varepsilon$:
\begin{align}\label{eqn:snr}
&\mathop{\mathbb{E}}_{\textbf{x} \sim p_\sigma}\left[\left\|\frac{\eta \nabla_\textbf{x}\log p_{\sigma,\gamma}(\textbf{x}|\textbf{m})}{\sqrt{2\eta}}\right\|^2\right]\notag\\
&\quad= \frac{\eta}{4}\mathop{\mathbb{E}}_{\textbf{x} \sim p_\sigma}\left[\left\|\nabla_\textbf{x}\log p_\gamma(\textbf{m}|\textbf{x}) + \nabla_\textbf{x}\log p_\sigma(\textbf{x})\right\|^2\right].
\end{align}
Assuming that gradients w.r.t. to the likelihood and the prior are uncorrelated, the SNR is approximately
\begin{equation}
\frac{\eta}{4}\mathop{\mathbb{E}}_{\textbf{x} \sim p_\sigma}\left[\left\|\nabla_\textbf{x}\log p_\gamma(\textbf{m}|\textbf{x})\right\|^2\right] + \frac{\eta}{4}\mathop{\mathbb{E}}_{\textbf{x} \sim p_\sigma}\left[\left\|\nabla_\textbf{x}\log p_\sigma(\textbf{x})\right\|^2\right].\
\end{equation}

Observe that $\log p_\gamma(\textbf{m}|\textbf{x})$ is a concave quadratic with smoothness proportional to $1/\gamma^2$; it follows analytically that $\mathbb{E}\left[\left\|\nabla_\textbf{x}\log p_\gamma(\textbf{m}|\textbf{x})\right\|^2\right] \propto 1/\gamma^2$. \citet{song2019generative} found empirically that $\mathbb{E} \|\nabla_\textbf{x}\log p_\sigma(\textbf{x})\|^2 \propto 1/\sigma^2$ for the NCSN model; we observe similar behavior for the flow-based Glow model \cite{kingma2018glow} and in Section \ref{sec:gradients} we propose a possible explanation for this behavior.
Therefore, to maintain a constant SNR, it suffices to set both $\gamma^2$ and $\sigma^2$ proportional to $\eta$.

\subsection{The gradients of the noisy prior}\label{sec:gradients}

We remark that the empirical finding $\mathbb{E} \|\nabla_\textbf{x}\log p_\sigma(\textbf{x})\|^2 \propto 1/\sigma^2$ discussed in Section \ref{sec:acceleration}, and the consistency of this observation across models and datasets, could be surprising. Gradients of the noisy densities $p_\sigma$ can be described by convolution of $p$ with a Gaussian kernel:
\begin{equation}
\nabla_\textbf{x} \log p_\sigma(\textbf{x}) = \nabla_\textbf{x}\log \mathop{\mathbb{E}}_{\epsilon \sim \mathcal{N}(0,I)}\left[p(\textbf{x} - \sigma\epsilon)\right].
\end{equation}
From this expression, assuming $p$ is continuous, we clearly see that the gradients are asymptotically independent of $\sigma$:
\begin{equation}
\lim_{\sigma \to 0} \nabla_\textbf{x} \log p_\sigma(\textbf{x}) = \nabla_\textbf{x}\log p(\textbf{x}).
\end{equation}
Maintaining proportionality $\mathbb{E} \|\nabla_\textbf{x}\log p_\sigma(\textbf{x})\|^2 \propto 1/\sigma^2$ requires the gradients to grow unbounded as $\sigma \to 0$, but the gradients of the noiseless distribution $\log p(\textbf{x})$ are finite. Therefore, proportionality must break down asymptotically and we conclude that--even though we turn the noise $\sigma^2$ down to visually imperceptible levels--we have not reached the asymptotic regime.

\begin{figure}
\centering
\includegraphics[scale=0.4]{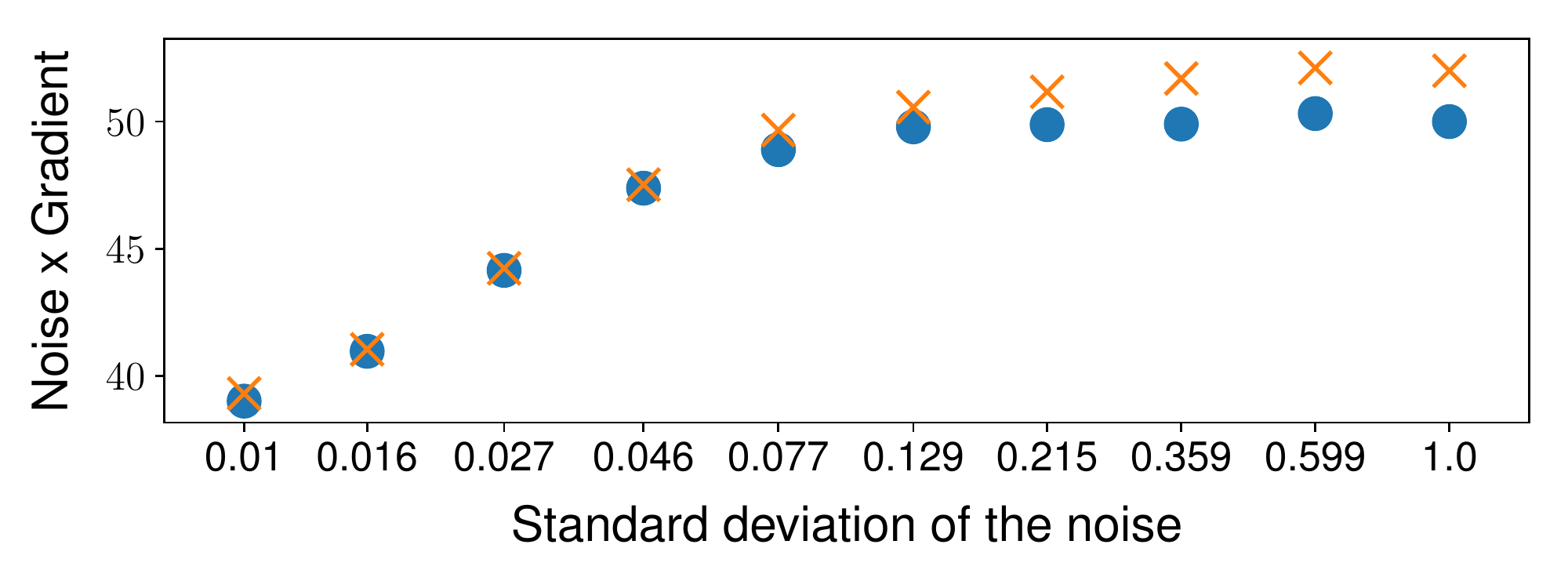}
\vspace*{-3mm}
\caption{The behavior of $\sigma \times \|\nabla_\textbf{x} \log p_\sigma(\textbf{x})\|$ in expectation for the NCSN (orange) and Glow (blue) models trained on CIFAR-10 at each of $10$ noise levels as $\sigma$ decays geometrically from $1.0$ to $0.01$. For large $\sigma$, $\|\nabla_\textbf{x} \log p_\sigma(\textbf{x})\| \approx 50/\sigma$. This proportional relationship breaks down for smaller $\sigma$. Because the expected gradient of the noiseless density $\log p(\textbf{x})$ is finite, its product with $\sigma$ must asymptotically approach zero as $\sigma \to 0$.}
\label{fig:proportionality}
\end{figure}

\begin{figure*}[h]
\centering
\hspace*{-2mm}\includegraphics[scale=0.8]{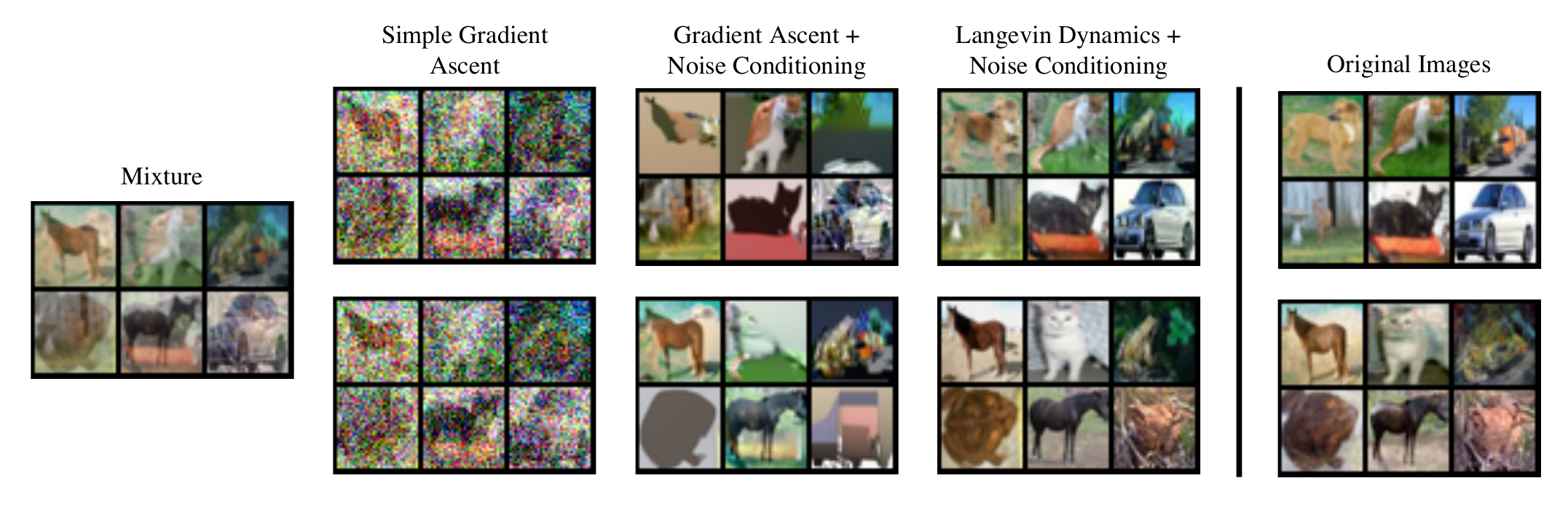}
\vspace*{-7mm}
\caption{Non-stochastic gradient ascent produces sub-par results. Annealing over smoothed-out distributions (Noise Conditioning) guides the optimization towards likely regions of pixel space, but gets stuck at sub-optimal solutions. Adding Gaussian noise to the gradients (Langevin dynamics) shakes the optimization trajectory out of bad local optima.}
\label{fig:algorithms}
\end{figure*}

We conjecture that the proportionality between the gradients and the noise is a consequence of severe non-smoothness in the noiseless model $p(\textbf{x})$. The probability mass of this distribution is peaked around plausible images $\textbf{x}$, and decays rapidly away from these points in most directions.  Consider the extreme case where the prior has a Dirac delta point mass. The convolution of a Dirac delta with a Gaussian is itself Gaussian so, near the point mass, the noisy distribution $p_\sigma$ will be proportional to a Gaussian density with variance $\sigma^2$. If $p_\sigma$ were exactly Gaussian then analytically
\begin{align}
&\mathop{\mathbb{E}}_{\textbf{x} \sim p_\sigma} \left[\|\nabla_\textbf{x} \log p_\sigma(\textbf{x})\|^2\right] = \frac{1}{\sigma^4}\mathop{\mathbb{E}}_{\textbf{x} \sim p_\sigma}\left[\textbf{x}^2\right]= \frac{1}{\sigma^2}.
\end{align}
Because the distribution $p(\textbf{x})$ does not contain actual delta spikes--only approximations thereof--we would expect this proportionality to eventually break down as $\sigma \to 0$. Indeed, Figure \ref{fig:proportionality} shows that both for NCSN and Glow models of CIFAR-10, after maintaining a very consistent proportionality $\mathbb{E} \left[\|\nabla_\textbf{x} \log p_\sigma(\textbf{x})\|^2\right] \propto 1/\sigma^2$ at the higher noise levels, the decay of $\sigma^2$ to zero eventually outpaces the growth of the gradients. 

\subsection{Hyper-parameter settings}\label{sec:hyper}

We adopt the hyper-parameters proposed by \citet{song2019generative} for annealing $\sigma^2$, the proportionality constant $\delta$, and the iteration count $T$. The noise $\sigma$ is geometrically annealed from $\sigma_1~=~1.0$ to $\sigma_{L}~=~0.01$ with $L=10$. We set $\delta = 2\times 10^{-5}$, and $T = 100$. We find that the same proportionality constant between $\sigma^2$ and $\eta$ also works well for $\gamma^2$ and $\eta$, allowing us to set $\gamma^2 = \sigma^2$. We use these hyper-parameters for both the NCSN and Glow models, applied to each of the three datasets MNIST, CIFAR-10, and LSUN.

\subsection{Constructing noise-conditioned models}\label{sec:models}

For noise-conditioned score networks, we can directly compute $\nabla_\textbf{x}\log p_\sigma(\textbf{x})$ by evaluating the score network at the desired noise level. For generative flow models like Glow, these noisy distributions are not directly accessible. We could estimate the distributions $p_\sigma(\textbf{x})$ by training Glow from scratch on datasets perturbed by each of the required noise levels $\sigma^2$. But this not practical; Glow is expensive to train, requiring thousands of epochs to converge and consuming hundreds of gpu-hours to obtain good models even for small low-resolution datasets.

Instead of training models $p_\sigma(\textbf{x})$ from scratch, we apply the concept of fine-tuning from transfer learning \cite{yosinski2014transferable}. Using pre-trained models of $p(\textbf{x})$ published by the Glow authors, we fine-tune these models on noise-perturbed data $\textbf{x} + \epsilon$, where $\epsilon \sim \mathcal{N}(0,\sigma^2I)$. Empirically, this procedure quickly converges to an estimate of $p_\sigma(\textbf{x})$, within about 10 epochs.

\subsection{The importance of stochasticity}\label{sec:stochasticity}

We remark that adding Gaussian noise to the gradients in the BASIS algorithm is essential. If we set aside the Bayesian perspective, it is tempting to simply run gradient ascent on the pixels of the components to maximize the likelihood of these components under the prior, with a Lagrangian term to enforce the mixture constraint $g(\textbf{x}) = \textbf{m}$:
\begin{equation}\label{eqn:gradasc}
\textbf{x} \gets \textbf{x} + \eta \nabla_\textbf{x} \left[\log p(\textbf{x}) - \lambda\|g(\textbf{x}) - \textbf{m}\|^2\right].
\end{equation}
But this does not work. As demonstrated in Figure \ref{fig:algorithms}, there are many local optima in the loss surface of $p(\textbf{x})$ and a greedy ascent procedure simply gets stuck. Pragmatically, the noise term in Langevin dynamics can be seen as a way to knock the greedy optimization \eqref{eqn:gradasc} out of local maxima.

In the recent literature, pixel-space optimizations by following gradients $\nabla_\textbf{x}$ of some objective are perhaps associated more with adversarial examples than with desirable results \cite{goodfellow2014explaining,nguyen2015deep}. We note that there have been some successes of pixel-wise optimization in texture synthesis \cite{gatys2015texture} and style transfer \cite{gatys2016image}. But broadly speaking, pixel-space optimization procedures often seem to go wrong. We speculate that noisy optimizations \eqref{eqn:langevin_fast} on smoothed-out objectives like $p_\sigma$ could be a widely applicable method for making pixel-space optimizations more robust.

%% file: eval.tex
\vspace{2mm}
\section{Evaluation Methodology}\label{sec:eval}

Many previous works on source separation evaluate their results using peak signal-to noise ratio (PSNR) or structural similarity index (SSIM) \cite{wang2004image}. These metrics assume that the original sources are identifiable; in probabilistic terms, the true posterior distribution $p(\textbf{x}|\textbf{m})$ is presumed to have a unique global maximum achieved by the ground truth sources (up to permutation of the sources). Under the identifiability assumption, it is reasonable to measure the quality of a separation algorithm by comparing separated sources to ground truth mixture components. PSNR, for example, evaluates separations by computing the mean-squared distance between pixel values of the ground truth and separated sources on a logarithmic scale.

For CIFAR-10 source separation, the ground truth source components of a mixture are not identifiable. As evidence for this claim, we call the reader's attention to Figure \ref{fig:ambiguity}. For each mixture depicted in Figure \ref{fig:ambiguity}, we present separation results that sum to the mixture and (to our eyes) look plausibly like CIFAR-10 images. However, in each case the separated images exhibit high deviation from the ground truth. This phenomenon is not unusual; Figure \ref{fig:sampling} shows an un-curated collection of samples from $p(\textbf{x}|\textbf{m})$ using BASIS, illustrating a variety of plausible separation results for each given mixture. We will later see evidence again of non-identifiability in Figure~\ref{fig:coloring}. If we accept that the separations presented in Figures \ref{fig:ambiguity}, \ref{fig:sampling}, and \ref{fig:coloring} are reasonable, then source separation on this dataset is fundamentally underdetermined; we cannot measure success using metrics like PSNR that compare separation results to ground truth.

Instead of comparing separations to ground truth, we propose instead to quantify the extent to which the results of a source separation algorithm look like samples from the data distribution. If a pair of images sum to the given mixture and look like samples from the data distribution, we deem the separation to be a success. This shift in perspective from identifiability of the latent components to the quality of the separated components is analogous to the classical distinction in the statistical literature between estimation and prediction \cite{shmueli2010explain,bellec2018slope}. To this end, we borrow the Inception Score (IS) \cite{salimans2016improved} and Frechet Inception Distance (FID) \cite{heusel2017gans} metrics from the generative modeling literature to evaluate CIFAR-10 separation results. These metrics attempt to quantify the similarity between two distributions given samples. We use them to compare the distribution of components produced by a separation algorithm to the distribution of ground truth images.

In contrast to CIFAR-10, the posterior distribution $p(\textbf{x}|\textbf{m})$ for an MNIST model is demonstrably peaked. Moreover, BASIS is able to consistently identify these peaks. This constitutes a constructive proof that components of MNIST mixtures are identifiable, and therefore comparisons to the ground-truth components make sense. We report PSNR results for MNIST, which allows us to compare the results of BASIS to other recent work on MNIST image separation \cite{halperin2018neural,Kong2019SingleChannelSS}.

%% file: separation.tex
\section{Experiments}\label{sec:experiments}

We evaluate results of BASIS on 3 datasets: MNIST \cite{lecun1998gradient} CIFAR-10 \cite{Krizhevsky09learningmultiple} and LSUN \cite{yu15lsun}. For MNIST and CIFAR-10, we consider both NCSN \cite{song2019generative} and Glow \cite{kingma2018glow} models as priors, using pre-trained weights published by the authors of these models. For LSUN there is no pre-trained NCSN model, so we consider results only with Glow. For Glow, we fine-tune the weights of the pre-trained models to construct noisy models $p_\sigma$ using the procedure described in Section \ref{sec:models}. Code and instructions for reproducing these experiments is available online.\footnote{\fontsize{7.9}{7.9}\url{https://github.com/jthickstun/basis-separation}}

\begin{figure}
\centering
\includegraphics[scale=0.9]{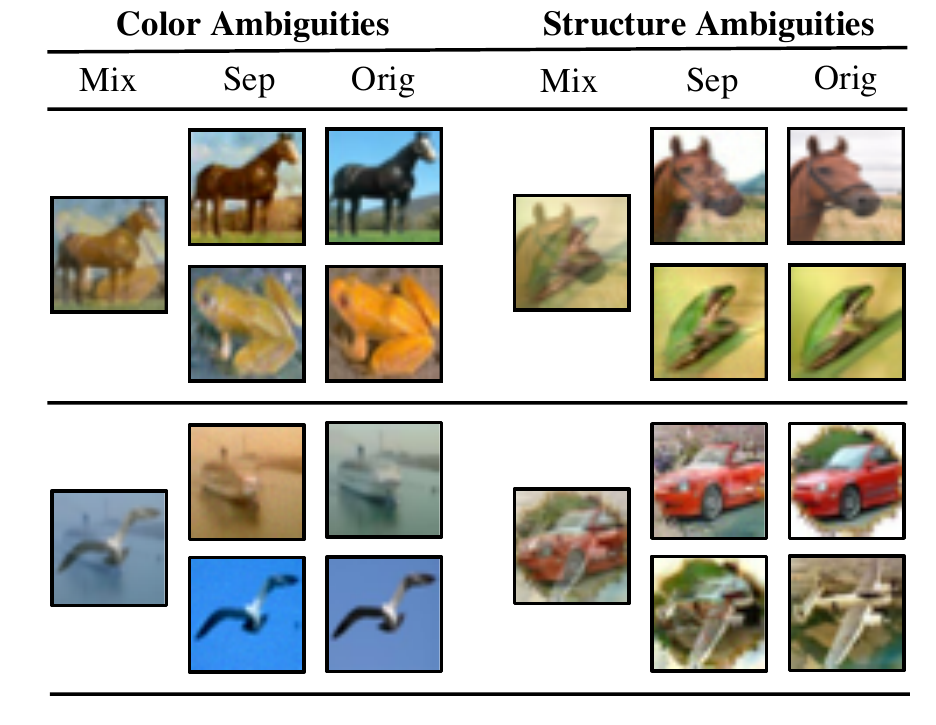}
\vspace*{-8mm}
\caption{A curated collection of examples demonstrating color and structural ambiguities in CIFAR-10 mixtures. In each case, the original components differ substantially from the components separated by BASIS using NCSN as a prior. But in each case, the separation results also look like plausible CIFAR-10 images.}
\label{fig:ambiguity}
\end{figure}

\begin{figure}
\centering
\includegraphics[scale=.7]{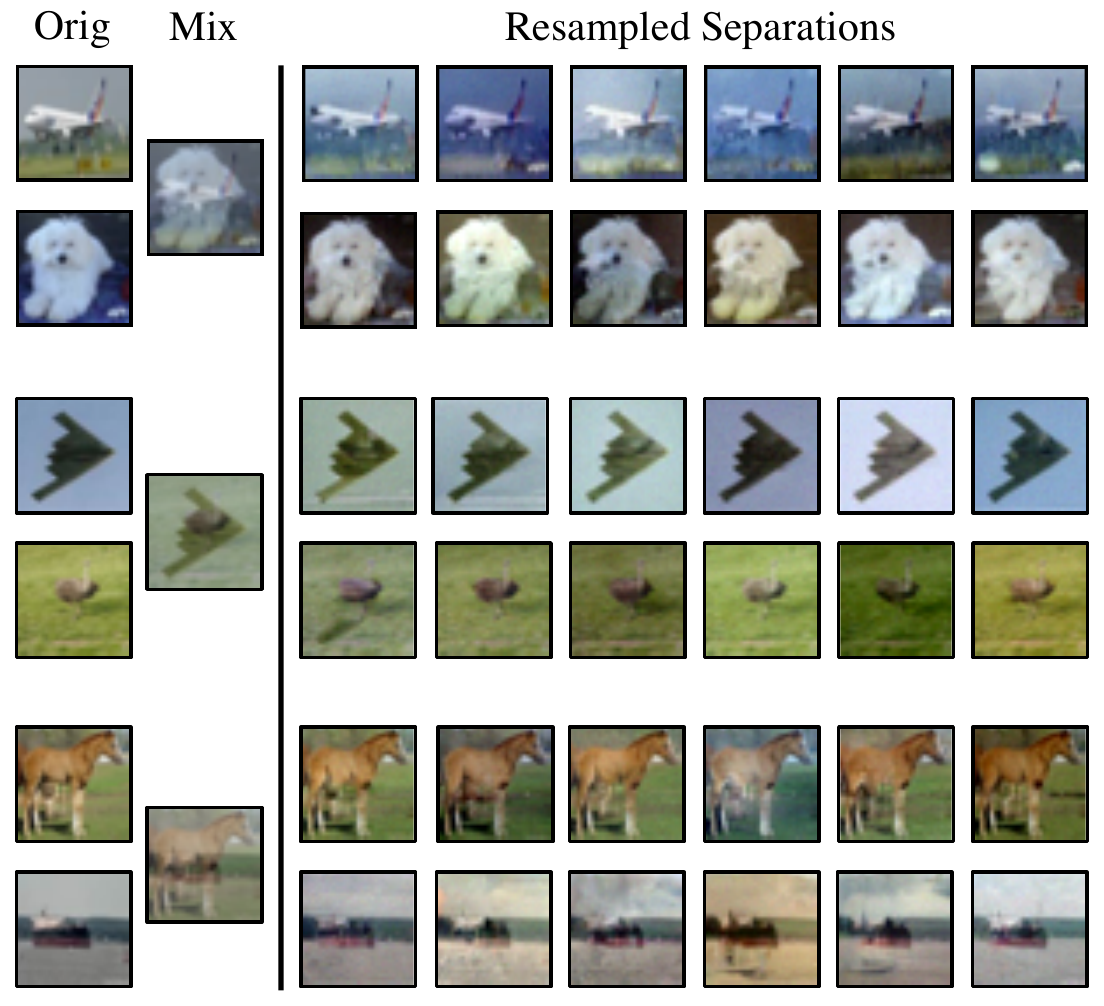}
\vspace*{-2mm}
\caption{Repeated sampling using BASIS with NCSN as a prior for several mixtures of CIFAR-10 images. While most separations look reasonable, variation in color and lighting makes comparative metrics like PSNR unreliable. This challenges the notion that the ground truth components are identifiable.}
\label{fig:sampling}
\end{figure}

\textbf{Baselines}. On MNIST we compare to results reported for the GAN-based ``S-D'' method \cite{Kong2019SingleChannelSS} and the fully supervised version of Neural Egg separation ``NES'' \cite{halperin2018neural}. Results for MNIST are presented in Section \ref{sec:mnist}. To the best of our knowledge there are no previously reported quantitative metrics for CIFAR-10 separation, so as a baseline we ran Neural Egg separation on CIFAR-10 using the authors' published code. CIFAR-10 results are presented in Section \ref{sec:cifar}. We present additional qualitative results for $64\times 64$ LSUN in Section \ref{sec:lsun}, which demonstrate that BASIS scales to larger images.

We also consider results for a simple baseline, ``Average,'' that separates a mixture $\textbf{m}$ into two 50\% masks $\textbf{x}_1 = \textbf{x}_2 = \textbf{m}/2$. This is a surprisingly competitive baseline. Observe that if we had no prior information about the distribution of components, and we measure separation quality by PSNR, then by a symmetry argument setting $\textbf{x}_1 = \textbf{x}_2$ is the optimal separation strategy in expectation. In principle we would expect Average to perform very poorly under IS/FID, because these metrics purport to measure similarity of distributions and mixtures should have little or no support under the data distribution. But we find that IS and FID both assign reasonably good scores to Average, presumably because mixtures exhibit many features that are well supported by the data distribution. This speaks to well-known difficulties in evaluating generative models \cite{theis2015note} and could explain the strength of ``Average'' as a baseline.

We remark that we cannot compare our algorithm to the separation-like task reported for CapsuleNets \cite{sabour2017dynamic}. The segmentation task discussed in that work is similar to source separation, but the mixtures used for the segmentation task are constructed using the non-linear threshold function $h(\textbf{x}) = \max(\textbf{x}_1 + \textbf{x}_2,1)$, in contrast to our linear function $g$. While extending the techniques of this paper to non-linear relationships between $\textbf{x}$ and $\textbf{m}$ is intriguing, we leave this to future this work.

\vspace{1mm}
\textbf{Class conditional separation}. The Neural Egg separation algorithm is designed with the assumption that the components $\textbf{x}_i$ are drawn from different distributions. For quantitative results on MNIST and CIFAR-10, we therefore consider two slightly different tasks. The first is class-agnostic, where we construct mixtures by summing randomly selected images from the test set. The second is class-conditional, where we partition the test set into two groupings: digits $0-4$ and $5-9$ for MNIST, animals and machines for CIFAR-10. The former task allows us compare to S-D results on MNIST, and the latter task allows us to compare to Neural Egg separation on MNIST and CIFAR-10.

There are two different ways to apply a prior for class-conditional separation. First observe that, because $\textbf{x}_1$ and $\textbf{x}_2$ are chosen independently,
\vspace*{-1mm}
\begin{equation}\label{eqn:factor}
p(\textbf{x}) = p(\textbf{x}_1,\textbf{x}_2) = p_1(\textbf{x}_1)p_2(\textbf{x}_2).
\vspace*{-1mm}
\end{equation}
In the class agnostic setting, $\textbf{x}_1$ and $\textbf{x}_2$ are drawn from the same distribution (the empirical distribution of the test set) so it makes sense to use a single prior $p = p_1 = p_2$. In the class conditional setting, we could potentially use separate priors over components $\textbf{x}_1$ and $\textbf{x}_2$. For the MNIST and CIFAR-10 experiments in this paper, we use pre-trained models trained on unconditional distribution of the training data for both the class agnostic and class conditional setting. It is possible that better results could be achieved in the class conditional setting by re-training the models on class conditional training data. For LSUN, the authors of Glow provide separate pre-trained models for the Church and Bedroom categories, so we are able to demonstrate class-conditional LSUN separations using distinct priors in Section \ref{sec:lsun}.

\textbf{Sample Likelihoods}. Although we do not directly model the posterior likelihood $p(\textbf{x}|\textbf{m})$, we can compute the log-likelihood of the output samples $\textbf{x}$. The log-likelihood is a function of the artificial variance hyper-parameter $\gamma$, so it is more informative to look at the unweighted square error $\|\textbf{m} - g(\textbf{x})\|^2$; this quantity can be interpreted as a reconstruction error, and measures how well we approximate the hard mixture constraint. Because we geometrically anneal the variance $\gamma$, by the end of optimization the mixture constraint is rigorously enforced; per-pixel reconstruction error is smaller than the quantization level of 8-bit color, resulting in pixel-perfect visual reconstructions.

For Glow, we can also compute the log-probability of samples under the prior. How do the probabilities of sources $\textbf{x}_\text{BASIS}$ constructed by BASIS separation compare to the probabilities of data $\textbf{x}_\text{test}$ taken directly from a dataset's test set? Because we anneal the noise to a fixed level $\sigma_L > 0$, we find it most informative to ask this question using the minimal-noise, fine-tuned prior $p_{\sigma_L}(\textbf{x})$. As seen in Table~\ref{tab:likelihood}, the outputs of BASIS separation are generally comparable in log-likelihood to test set images; BASIS separation recovers sources deemed typical by the prior.

\begin{table}[b]
\vspace{-2mm}
\centering 
\caption{The mean log-likelihood under the minimal-noise Glow prior $p_{\sigma_L}(\textbf{x})$ for the test set $\textbf{x}_\text{test}$, and for samples of 100 BASIS separations $\textbf{x}_\text{BASIS}$. The log-likelihood of each test set under the noiseless prior $p(\textbf{x}_\text{test})$ is reported for reference.}
\bigskip
\vspace{-1mm}
\label{tab:likelihood}
  \begin{tabular}{l c c c}
  \hline
  Dataset & $p(\textbf{x}_\text{test})$ & $p_{\sigma_L}(\textbf{x}_\text{test})$ & $p_{\sigma_L}(\textbf{x}_\text{BASIS})$ \\
  \hline
  MNIST & 0.5 & 3.6 & 3.6 \\
  CIFAR-10 & 3.4 & 4.5 & 4.7 \\
  LSUN (bed) & 2.4 & 4.2 & 4.4 \\
  LSUN (crh) & 2.7 & 4.4 & 4.4 \\
 \hline
  \end{tabular}
\end{table}

\subsection{MNIST separation}\label{sec:mnist}

\begin{figure}
\centering
\includegraphics[scale=0.41]{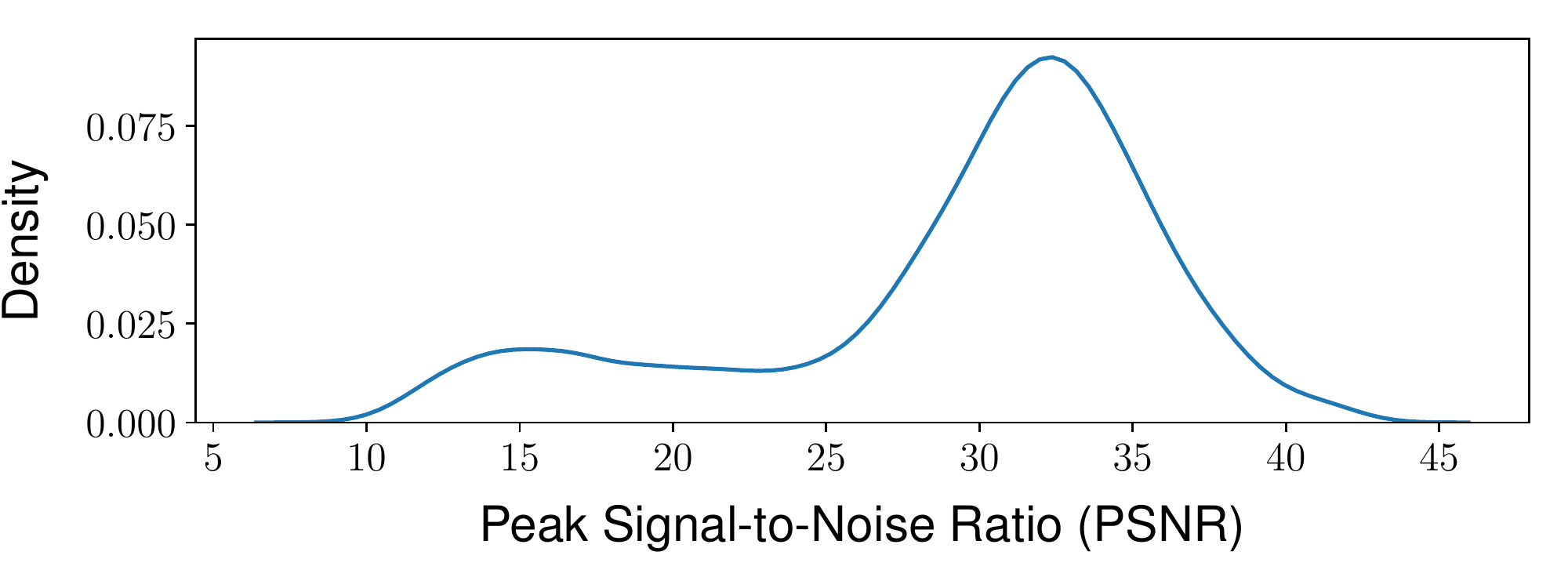}
\vspace{-8mm}
\caption{The empirical distribution of PSNR for 5,000 class agnostic MNIST digit separations using BASIS with the NCSN prior (see Table \ref{tab:mnist} for comparison of the central tendencies of this and other separation methods).}
\label{fig:distribution}
\end{figure}

Quantitative results for MNIST image separation are reported in Table \ref{tab:mnist}, and a panel of visual separation results are presented in Figure \ref{fig:teaser}. For quantitative results, we report mean PSNR over separations of $12,000$ separated components. The distribution of PSNR for class agnostic MNIST separation is visualized in Figure \ref{fig:distribution}. We observe that approximately $2/3$ of results exceed the mean PSNR of 29.5, which to our eyes is visually indistinguishable from ground truth.

\begin{table}[b!]
\centering 
\caption{PSNR results for separating 6,000 pairs of equally mixed MNIST images. For class split results, one image comes from label $0-4$ and the other comes from $5-9$. We compare to S-D \cite{Kong2019SingleChannelSS}, NES  \cite{halperin2018neural}, convolutional NMF (class split) \cite{halperin2018neural} and standard NMF (class agnostic)  \cite{Kong2019SingleChannelSS}.}
\bigskip
\label{tab:mnist}
  \begin{tabular}{l c c}
  \hline
  Algorithm & Class Split & Class Agnostic\\
  \hline
  Average & 14.8 & 14.9 \\
  NMF & 16.0 & 9.4\\
  S-D & - & 18.5 \\
  BASIS (Glow) & 22.9 & 22.7 \\
  NES& 24.3 & -\\
  BASIS (Glow, 10x) & 27.7 & 27.1 \\
  \textbf{BASIS (NCSN)} & \textbf{29.5} & \textbf{29.3}\\
 \hline
  \end{tabular}
\end{table}

\begin{figure*}[h]
\centering
\hspace*{-1mm}\includegraphics[scale=1.9]{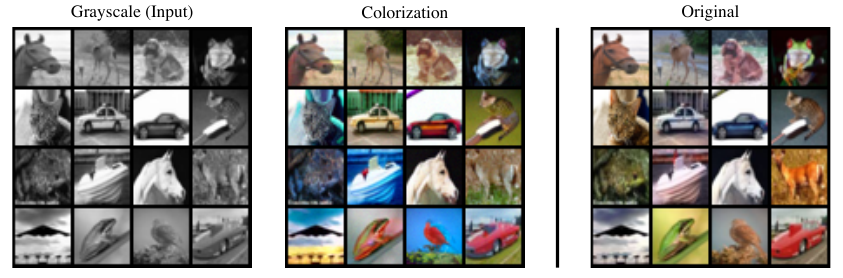}
\caption{Colorizing CIFAR-10 images. Left: original CIFAR-10 images. Middle: greyscale conversions of the images on the left. Right: imputed colors for the greyscale images, found by BASIS using NCSN as a prior.}
\label{fig:coloring}
\vspace{-3mm}
\end{figure*}

A natural approach to improve separation performance is to sample multiple $\textbf{x} \sim p(\cdot|\textbf{m})$ for a given mixture $\textbf{m}$. A major advantage of models like Glow, that explicitly parameterize the prior $p(\textbf{x})$, is that we can approximate the maximum of the posterior distribution with the maximum over multiple samples. By construction, samples from BASIS approximately satisfy $g(\textbf{x}) = \textbf{m}$, so for the noiseless model we simply declare $p(\textbf{m}|\textbf{x}) = 1$ and therefore $p(\textbf{x}|\textbf{m}) \propto p(\textbf{x})$. We demonstrate the effectiveness of resampling in Table~\ref{tab:mnist} (Glow, 10x) by comparing the expected PSNR of $\textbf{x} \sim p(\cdot|\textbf{m})$ to the expected PSNR of $\argmax_i p(\textbf{x}_i)$ over $10$ samples $\textbf{x}_1,\dots,\textbf{x}_{10} \sim p(\cdot|\textbf{m})$.  Even moderate resampling dramatically improves separation performance. Unfortunately this approach cannot be applied to the otherwise superior NCSN model, which does not model explicit likelihoods $p(\textbf{x})$. 

Without any modification, we can apply BASIS to separate mixtures of $k > 2$ images. We contrast this with regression-based methods, which require re-training to target varying numbers of components. Figure  \ref{fig:teaser} shows the results of BASIS using the NCSN prior applied to mixtures of four randomly selected images. For more mixture components, we observe that identifiability of ground truth sources begins to break down. This is illustrated by looking at the central item in each panel of Figure \ref{fig:teaser} (highlighted in orange).

\subsection{CIFAR-10}\label{sec:cifar}

\begin{table}[b!]
\centering 
\caption{Inception Score / FID Score of 25,000 separations (50,000 separated images) of two overlapping CIFAR-10 images using NCSN as a prior. In Class Split one image comes from the category of animals and other from the category of vehicles. NES results using published code from \citet{halperin2018neural}.}
\bigskip
\label{tab:cifar}
  \begin{tabular}{l c c}
  \hline
  Algorithm & Inception Score & FID\\
  \hline
  Class Split\\
  \hline
  \hline
  NES & 5.29 $\pm$ 0.08  & 51.39\\
  BASIS (Glow) & 5.74 $\pm$ 0.05 & 40.21 \\
  Average & 6.14 $\pm$ 0.11 & 39.49 \\
  \textbf{BASIS (NCSN)} & \textbf{7.83 $\pm$ 0.15} & \textbf{29.92}\\
 \hline
  Class Agnostic\\
  \hline
  \hline
  BASIS (Glow) & 6.10 $\pm$  0.07 & 37.09 \\
  Average & 7.18 $\pm$ 0.08 & 28.02 \\
  \textbf{BASIS (NCSN)} & \textbf{8.29 $\pm$ 0.16} & \textbf{22.12}\\
 \hline
  \end{tabular}
\end{table}

Quantitative results for CIFAR-10 image separation measured are presented in Table \ref{tab:cifar}, and visual separation results are presented in Figure \ref{fig:teaser}.

We can also view image colorization \cite{levin2004colorization,zhang2016colorful} as a source separation problem by interpreting a grayscale image as a mixture of the three color channels of an image $\textbf{x} = (\textbf{x}_r,\textbf{x}_g,\textbf{x}_b)$ with
\begin{equation}
g(\textbf{x}) = (\textbf{x}_r + \textbf{x}_g + \textbf{x}_b)/3.
\end{equation}
Unlike our previous separation problems, the channels of an image are clearly not independent, and the factorization of $p$ given by Equation \ref{eqn:factor} is unwarranted. But conveniently, a generative model trained on color CIFAR-10 images itself models the joint distribution $p(\textbf{x}) = p(\textbf{x}_r,\textbf{x}_g,\textbf{x}_b)$. Therefore, the same pre-trained generative model that we use to separate images can also be used to color them.

Qualitative colorization results are visualized in Figure \ref{fig:coloring}. The non-identifiability of ground truth is profound for this task (see Section~\ref{sec:eval} for discussion of identifiability). We draw attention to the two cars in the middle of the panel: the white car that is colored yellow by the algorithm, and the blue car that is colored red. The colors of these specific cars cannot be inferred from a greyscale image; the best an algorithm can do is to choose a reasonable color, based on prior information about the colors of cars.

Quantitative coloring results for CIFAR-10 are presented in Table \ref{tab:coloring}. We remark that the IS and FID scores for coloring are substantially better than the IS and FID scores of 8.87 and 25.32 respectively reported for unconditional samples from the NCSN model; conditioning on a greyscale image is enormously informative. Indeed, the Inception Score of NCSN-colorized CIFAR-10 is close to the Inception Score of the CIFAR-10 dataset itself.

\begin{table}
\centering 
\caption{Inception Score / FID Score of 50,000 colorized CIFAR-10 images. As measured by IS/FID, the quality of NCSN colorizations nearly matches CIFAR-10 itself.}
\bigskip
\vspace{1mm}
\label{tab:coloring}
  \begin{tabular}{l c c}
  \hline
  Data Distribution & Inception Score & FID Score\\
  \hline
  Input Grayscale & 8.01 $\pm$ 0.10 &  68.52 \\
  BASIS (Glow) & 8.69 $\pm$ 0.15 & 28.70 \\
  \textbf{BASIS (NCSN)} & \textbf{10.53} $\pm$ \textbf{0.17} & \textbf{11.58} \\
  CIFAR-10 Original& 11.24 $\pm$ 0.12 & 0.00 \\
 \hline
  \end{tabular}
  \vspace{2.5mm}
\end{table}

\subsection{LSUN separation}\label{sec:lsun}

\begin{figure}[h]
\centering
\vspace{-2mm}
\hspace*{-1mm}\includegraphics[scale=.32]{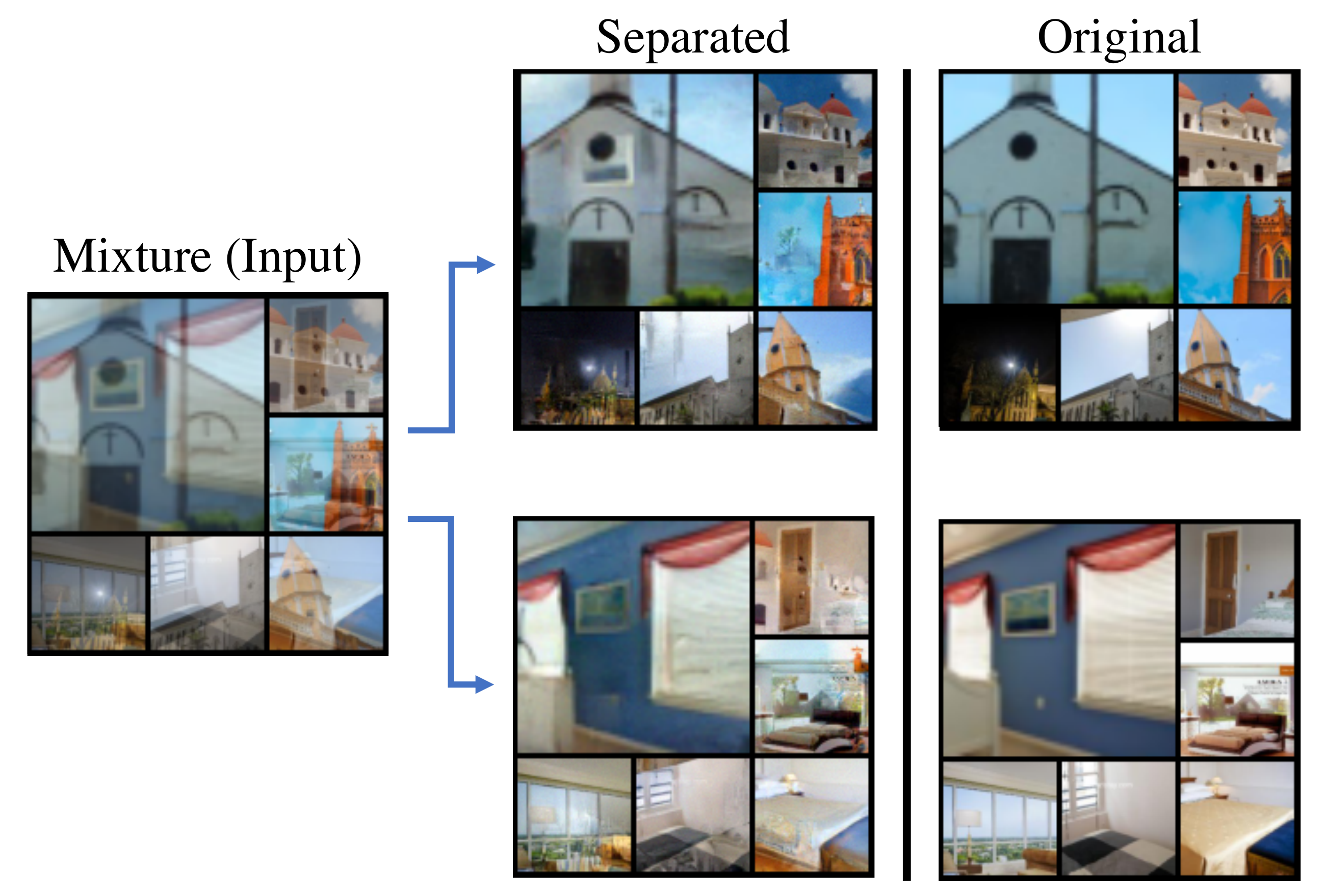}
\vspace{-8mm}
\caption{$64\times 64$ LSUN separation results using Glow as a prior. One mixture component is sampled from the LSUN churches category, and the other component is sampled from LSUN bedrooms.}
\label{fig:lsun}
\end{figure}

Qualitative results for LSUN separations are visualized in Figure \ref{fig:lsun}.
While the separation results in Figure \ref{fig:lsun} are imperfect, Table \ref{tab:likelihood} shows that the mean log-likelihood of the separated components is comparable to the mean log-likelihood that the model assigns to images in the test set. This suggests that the model is incapable of distinguishing these separations from better results, and the imperfections are attributable to the quality of the model rather than to the separation algorithm. This is encouraging, because it suggests that the artifacts are due to the Glow model rather than the BASIS separation algorithm, and that better separation results will be achievable with improved generative models.

%% file: conclude.tex
\section{Conclusion}

In this paper, we introduced a new approach to source separation that makes use of a likelihood-based generative model as a prior. We demonstrated the ability to swap in different generative models for this purpose, presenting results of our algorithm using both NCSN and Glow. We proposed new methodology for evaluating source separation on richer datasets, demonstrating strong performance on MNIST and CIFAR-10. Finally, we presented qualitative results on LSUN that point the way towards scaling this method to practical tasks such as speech separation, using generative audio models like WaveNets \cite{oord2016wavenet}.

%% file: appendix.tex
\section{Experimental Details}

\subsection{Fine-tuning}

We fine-tuned the MNIST, CIFAR-10, and Glow models at 10 noise levels $\sigma^2$ (see Section \ref{sec:hyper}) for 50 epochs each on clusters of 4 1080Ti GPU's. This procedure converges rapidly, with no further decrease of the negative log-likelihood after the first 10 epochs. Although Glow models theoretically have full support, the noiseless pre-trained models assign vanishing probability to highly noisy images. In practice, this can cause invertibility assertion failures when fine-tuning directly from the noiseless model. To avoid this we took an iterative approach: first fine-tune the lowest noise level $\sigma = .01$ from the noiseless model, then fine-tune the $\sigma = .016$ model from the $\sigma = .01$ model, etc.

\subsection{Scaling and Resources}

Scaling Algorithm~\ref{alg:sourcesep} to richer datasets is constrained primarily by the limited availability of strong, likelihood-based generative models for these datasets. For high resolution images, the running time of Algorithm \ref{alg:sourcesep} can also become substantial. Assuming the hyper-parameters $T$ and $L$ discussed in Section \ref{sec:hyper} remain valid at higher resolutions, the computational complexity of BASIS scales linearly with the cost of evaluating gradients of the model (albeit with a large multiplicative constant $T\times L$). Therefore, if a generative model is tractable to train, then it should also be tractable to use for BASIS separation.

In concrete detail, we observe that a batch of $50$ BASIS separation results for MNIST or CIFAR-10 using NCSN takes  $< 5$ minutes on a single 1080Ti GPU. Running BASIS with Glow is much slower. We observe that substantial time is spent loading and unloading the noisy models $p_\sigma$ from memory (in contrast to NCSN, which uses a single noise-conditioned model).  A batch of $50$ BASIS separation results on MNIST or CIFAR-10 using Glow takes about 30 minutes on a 1080Ti. A batch of $9$ BASIS separation result on LSUN using Glow takes 2-3 hours on a 1080Ti.

\subsection{Visual Comparisons}

When using class-agnostic priors, BASIS separation is symmetric in its output components. To facilitate visual comparisons between original images and separated components, we sort the BASIS separated components to minimize PSNR to the original images. This usually results in the separated components being visually paired with the most similar original components. But due to the deficiencies of PSNR as a comparative metric this is not always the case; the alert reader may have noticed that the yellow and silver car mixture in Figure \ref{fig:teaser} appears to have been displayed in reverse order. This happens because the separated yellow car component takes the light sky from the original silver car component, and the lightness of the sky dominates the PSNR metric.

For the LSUN separation results, where we use a church model for the first component and a bedroom model for the second, the symmetry is broken. For these results, components naturally sort themselves into church and bedroom components, which can be compared directly to the original images.

\clearpage

\section{Intermediate Samples During the Annealing Process}

\begin{figure}[h!]
\centering
\hspace*{18mm}\includegraphics[scale=.9]{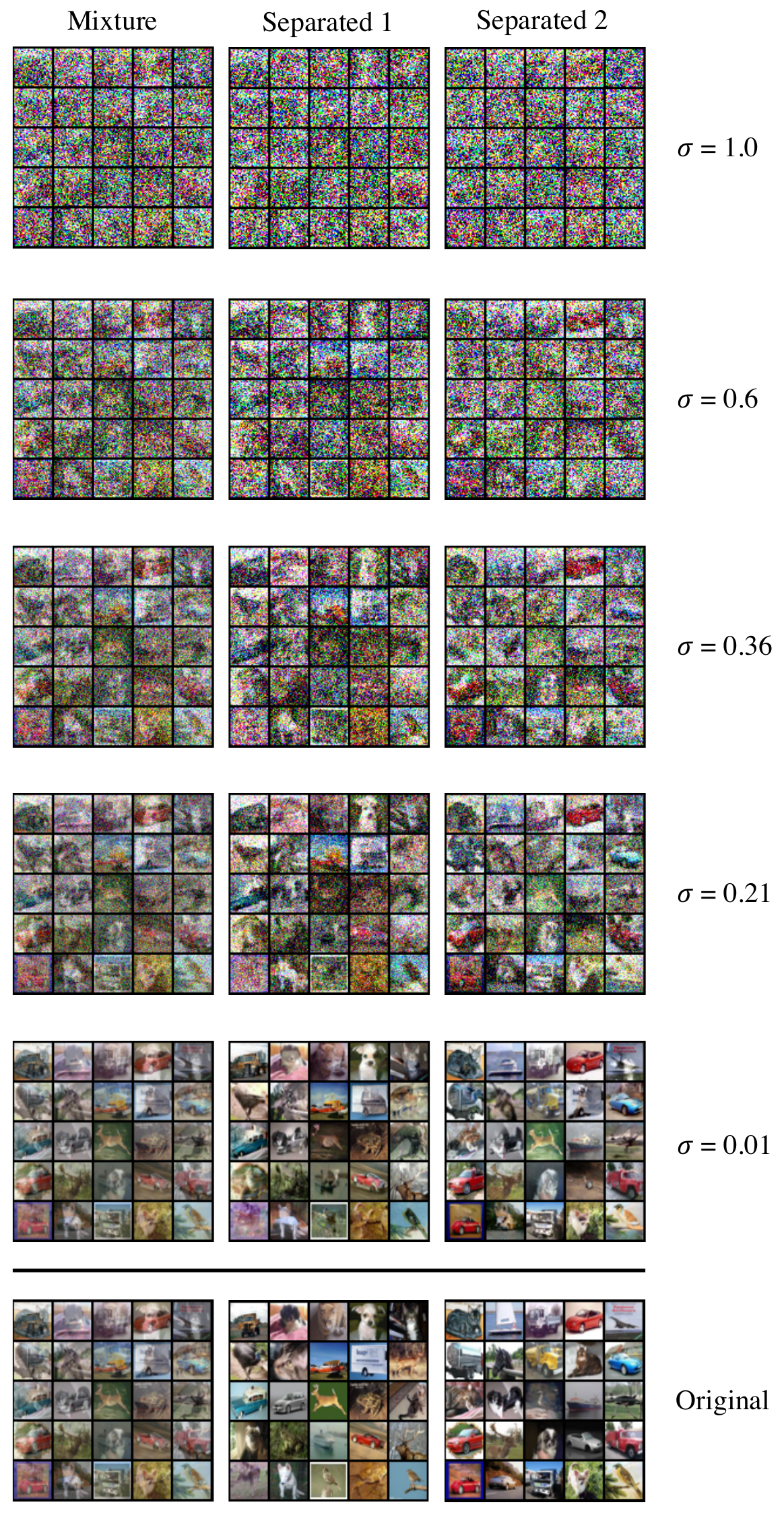}
\caption{Intermediate CIFAR-10 separation results taken at noise levels $\sigma$ during the annealing process of BASIS separation.}
\end{figure}

\clearpage

\section{MNIST Separation Results Under Different Models and Sampling Procedures}

\begin{figure}[h!]
\centering
\vspace{3cm}
\includegraphics[scale=1.0]{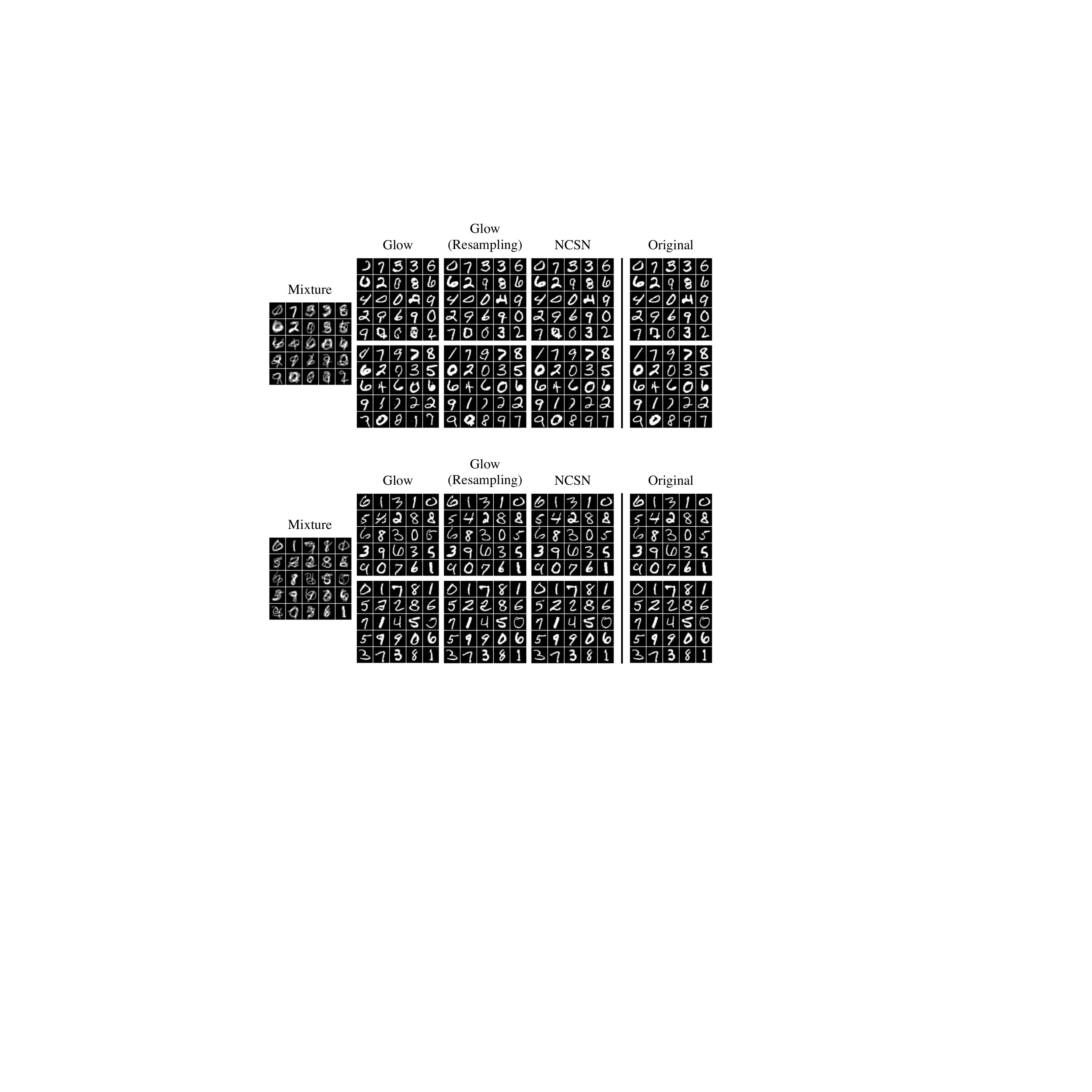}
\caption{Uncurated class-agnostic separation results using: (1) samples from the posterior with Glow as a prior (2) an approximate MAP estimate using the maximum over 10 samples from the posterior with Glow as a prior (3) samples from the posterior with NCSN as a prior.}
\end{figure}

\clearpage

\section{Extended CIFAR-10 Separation Results}

\subsection{NCSN Prior}

\begin{figure}[h!]
\centering
\includegraphics[scale=1.30]{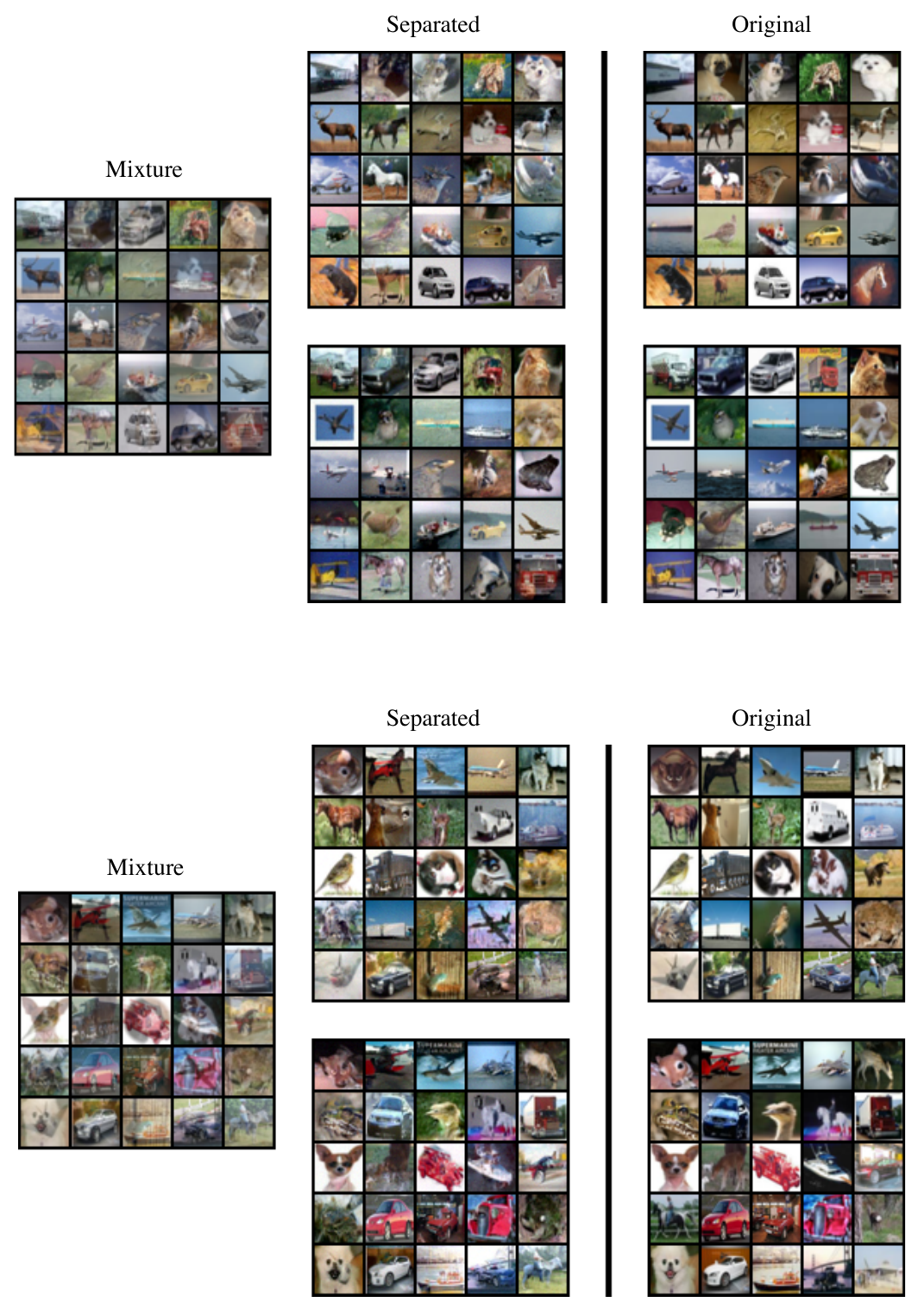}
\caption{Uncurated class-agnostic CIFAR-10 separation results using NCSN as a prior.}
\end{figure}

\clearpage

\subsection{Glow Prior}

\begin{figure}[h!]
\centering
\includegraphics[scale=1.30]{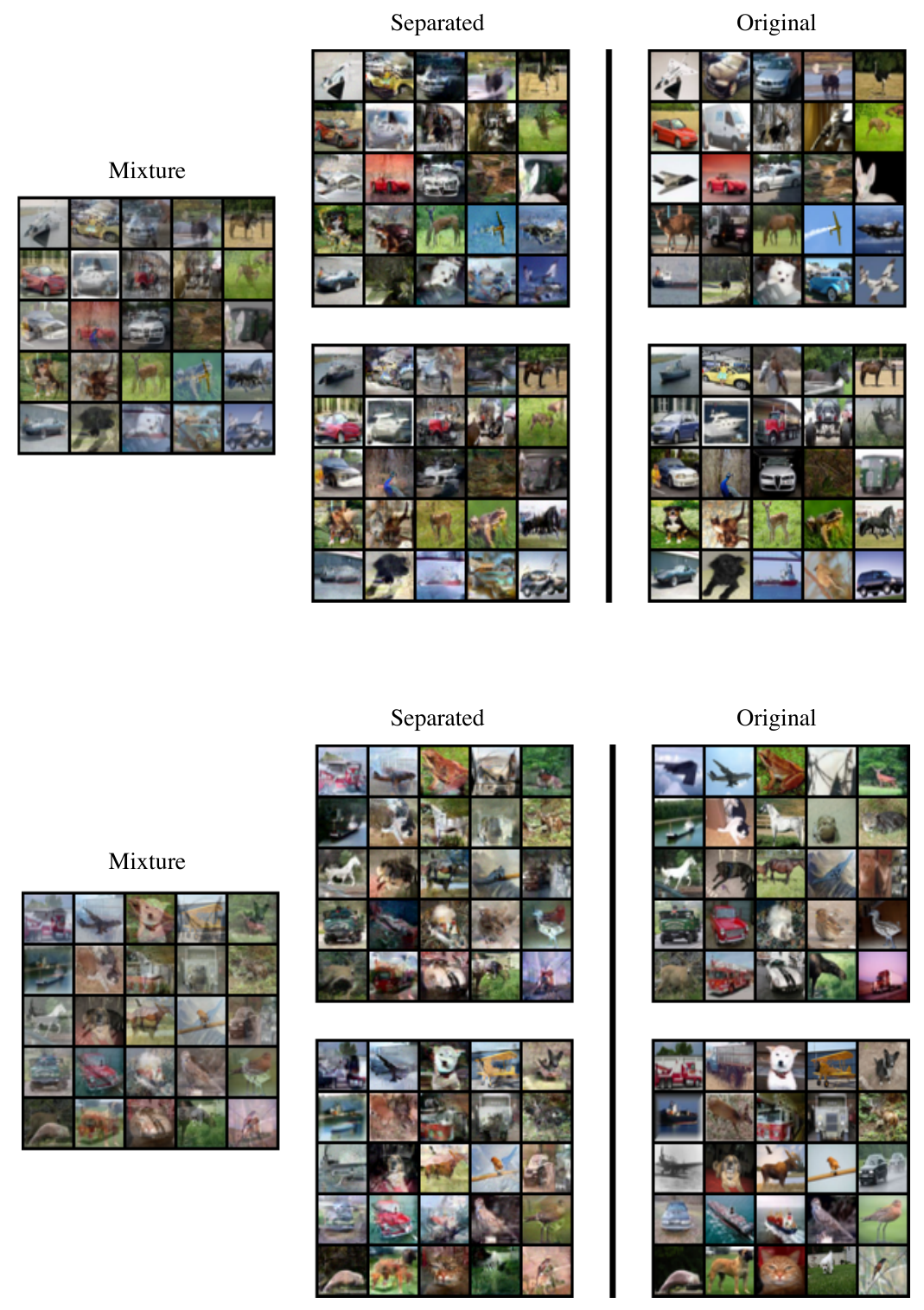}
\caption{Uncurated class-agnostic CIFAR-10 separation results using Glow as a prior.}
\end{figure}

\clearpage

\section{Extended CIFAR-10 Colorization Results}

\subsection{NCSN Prior}

\begin{figure}[h!]
\centering
\includegraphics[scale=1.35]{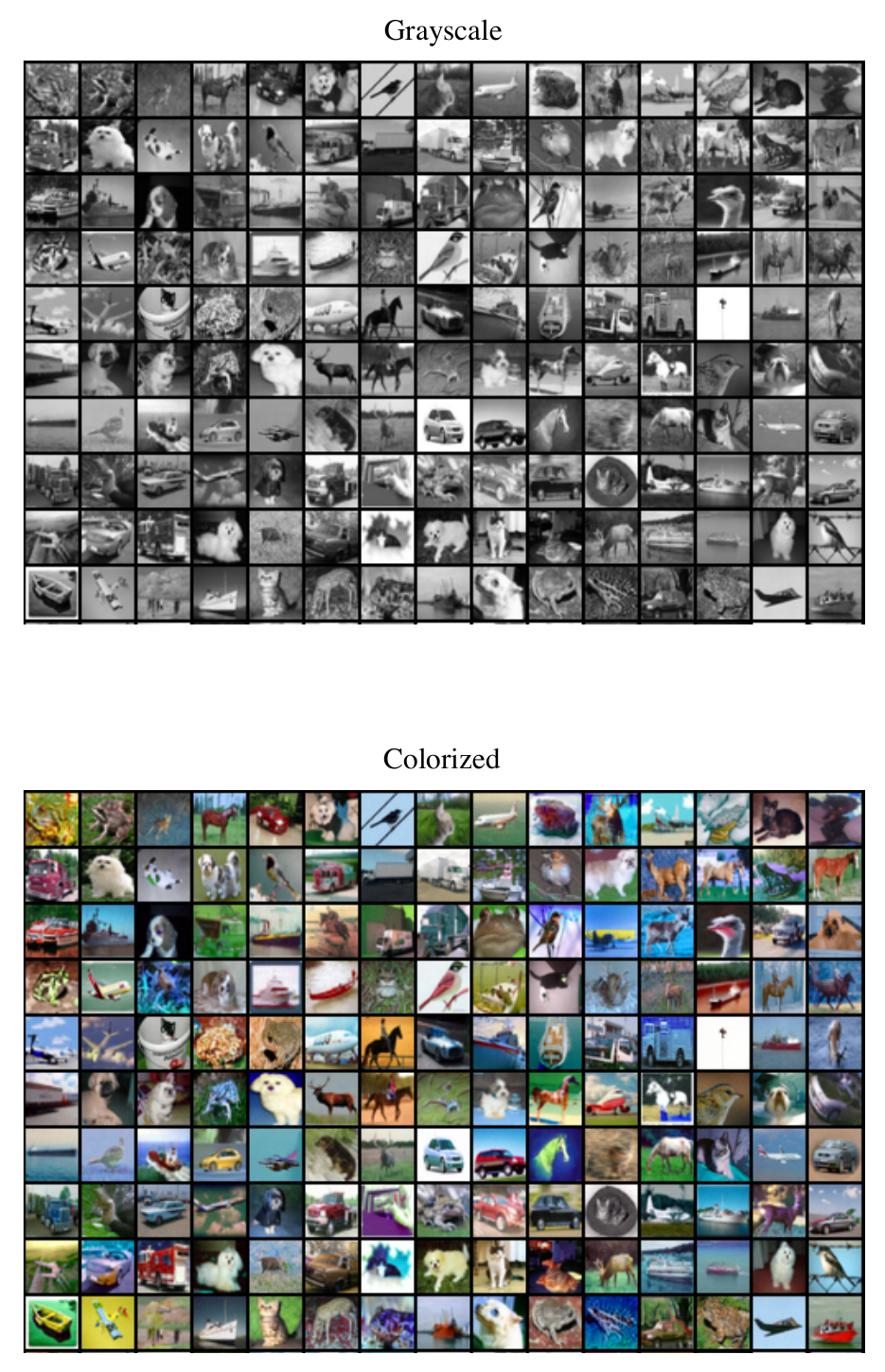}
\caption{Uncurated CIFAR-10 colorization results using NCSN as a prior.}
\end{figure}

\clearpage

\subsection{Glow Prior}

\begin{figure}[h!]
\centering
\includegraphics[scale=1.35]{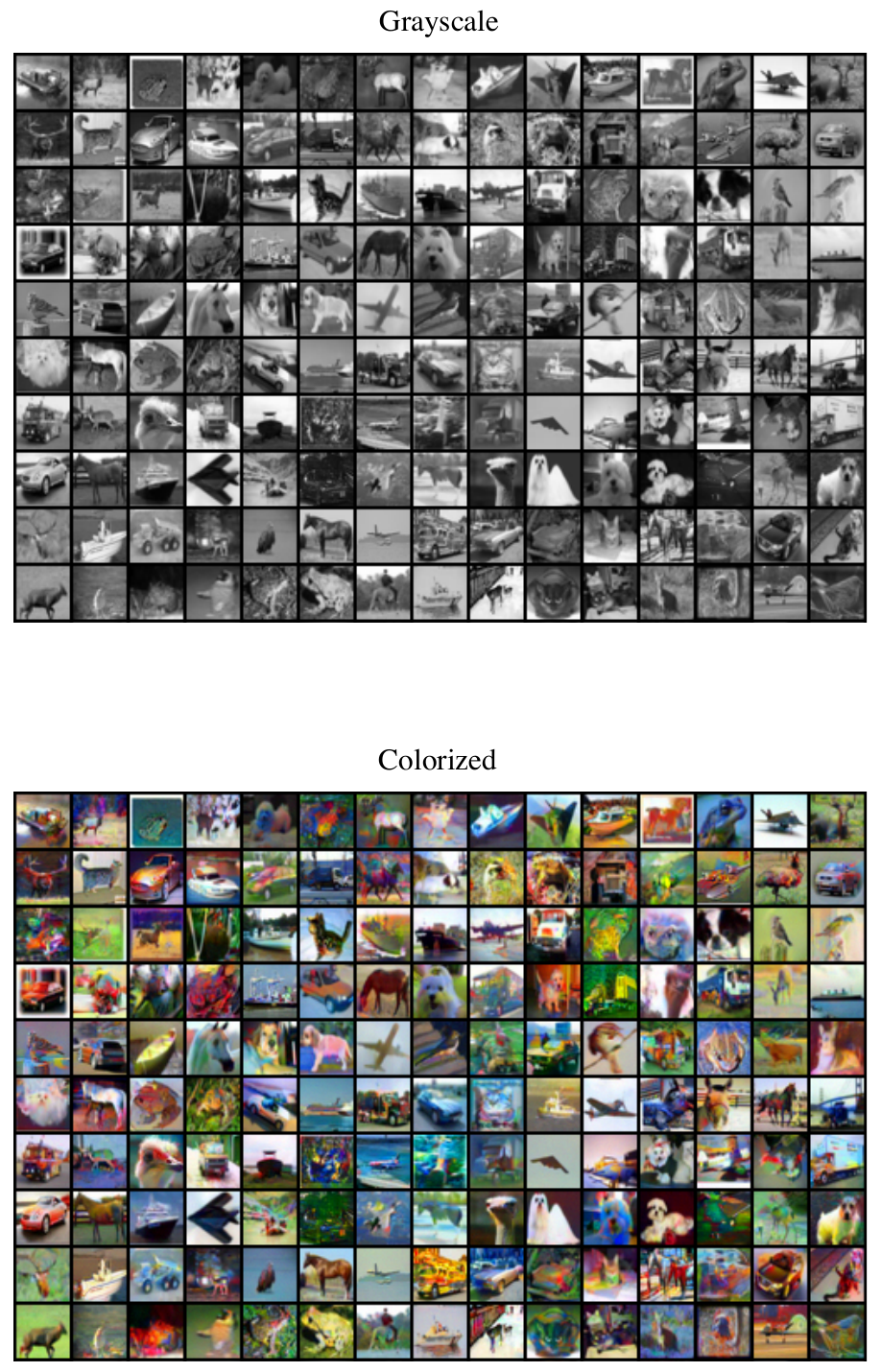}
\caption{Uncurated CIFAR-10 colorization results using Glow as a prior.}
\end{figure}

\clearpage

\section{Extended LSUN Separation Results}

\begin{figure}[h!]
\centering
\includegraphics[scale=1.1]{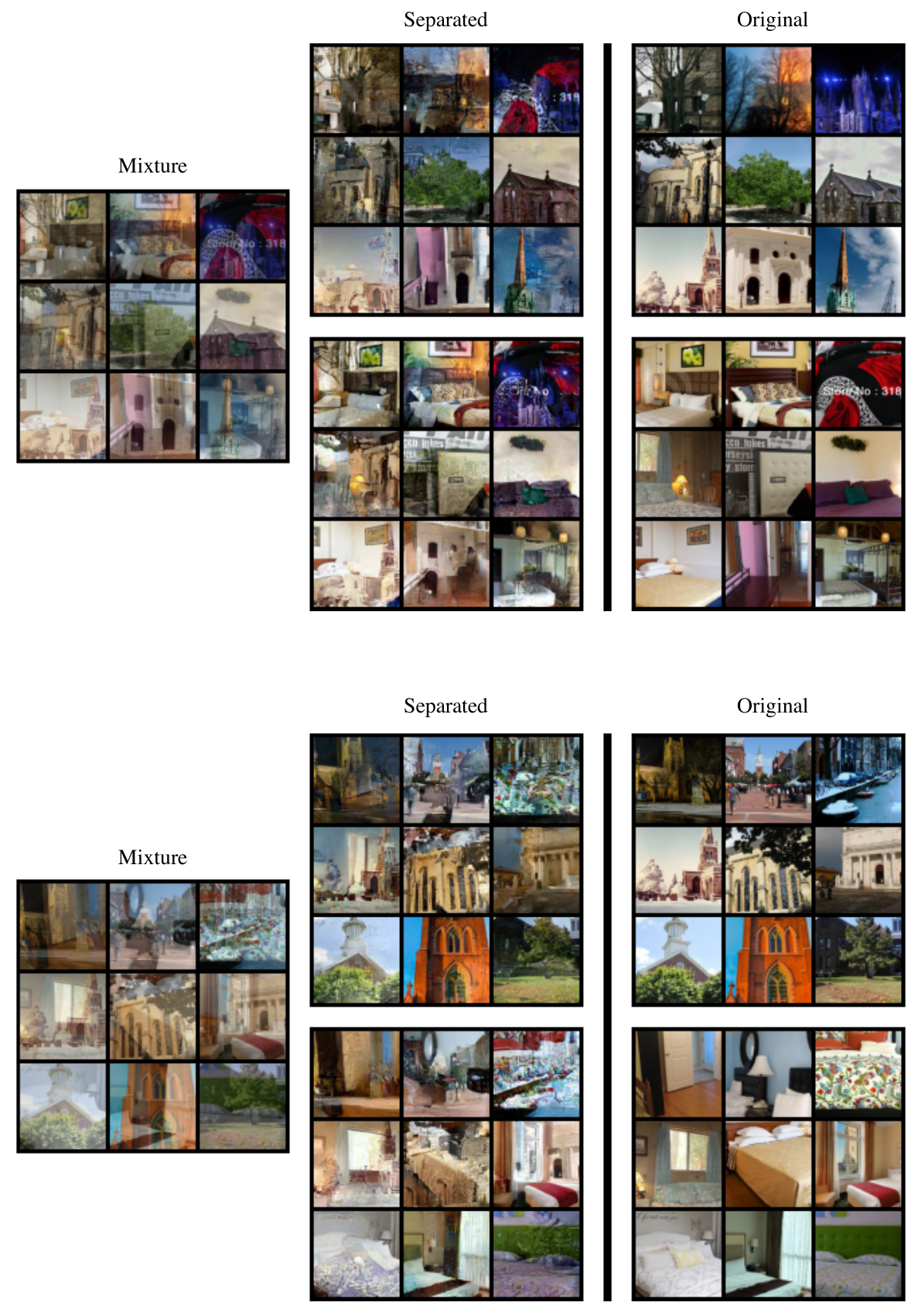}
\caption{Uncurated church/bedroom LSUN separation results using Glow as a prior.}
\end{figure}